\documentclass[letterpaper,10pt,conference]{ieeeconf}

\IEEEoverridecommandlockouts

\usepackage[utf8]{inputenc}                 
\usepackage[T1]{fontenc}                    
\usepackage[colorlinks=true]{hyperref}      
\usepackage{url}                            
\usepackage{booktabs}                       
\usepackage{amsfonts}                       
\usepackage{nicefrac}                       
\usepackage{microtype}                      
\usepackage{xcolor}
\usepackage{amsfonts,amsmath,amssymb}
\usepackage{bbm}
\usepackage[ruled,vlined]{algorithm2e}
\usepackage{thmtools, thm-restate}
\usepackage{graphicx}
\usepackage{cite}
\usepackage{mathtools}
\usepackage{amssymb}

\usepackage{hyperref}
\usepackage{footmisc}
\usepackage{bm}
\usepackage{bibentry}
\usepackage{url}
\usepackage{graphicx}
\usepackage{multicol}
\usepackage{amsmath}

\usepackage{amsthm}

\usepackage[capitalise,nameinlink]{cleveref}    
\crefformat{equation}{#2(#1)#3}
\crefformat{figure}{#2Fig.~#1#3}
\crefformat{section}{#2\S#1#3}
\crefformat{subsection}{#2\S#1#3}
\crefformat{subsubsection}{#2\S#1#3}
\crefformat{assumption}{#2Assumption~#1#3}

\newcommand{\iid}{\overset{\textup{iid}}{\sim}}

\newcommand{\x}{\bm{x}}
\newcommand{\y}{\bm{y}}
\newcommand{\traindata}{\mathcal{D}_\mathrm{train}}
\newcommand{\testdata}{\mathcal{D}_\mathrm{test}}
\newcommand{\ndata}{N}
\newcommand{\ptrain}{p_\mathrm{train}}
\newcommand{\ptest}{p_\mathrm{test}}
\newcommand{\learnedmodel}{f}

\newcommand{\Ptrain}{P_\mathrm{train}}
\newcommand{\Ptest}{P_\mathrm{test}}

\newcommand{\RQ}[1]{\textbf{RQ #1}}

\title{\LARGE \bf%
A System-Level View on Out-of-Distribution Data in Robotics
}
\author{Rohan Sinha, Apoorva Sharma, Somrita Banerjee, Thomas Lew, Rachel Luo, \\ Spencer~M.~Richards, Yixiao Sun, Edward Schmerling, Marco Pavone
\thanks{%
    The authors are with the Autonomous Systems Lab at Stanford University, Stanford, CA. \texttt{\{rhnsinha, apoorva, somrita, thomas.lew, rsluo, 
 spenrich, alvinsun, schmrlng, pavone\}@stanford.edu\}}
}%
}

\newtheoremstyle{rqstyle}
  {2pt} 
  {2pt} 
  {} 
  {} 
  {\bfseries} 
  {.} 
  {.5em} 
  {} 

\theoremstyle{rqstyle}
\newtheorem{researchquestion}{RQ}

\renewcommand{\baselinestretch}{0.993}
\begin{document}

\maketitle
\thispagestyle{empty}
\pagestyle{empty}

\begin{abstract}
When testing conditions differ from those represented in training data, so-called out-of-distribution (OOD) inputs can mar the reliability of learned components in the modern robot autonomy stack. Therefore, coping with OOD data is an important challenge on the path towards trustworthy learning-enabled open-world autonomy. In this paper, we aim to demystify the topic of OOD data and its associated challenges in the context of data-driven robotic systems, drawing connections to emerging paradigms in the ML community that study the effect of OOD data on learned models in isolation. We argue that as roboticists, we should reason about the overall \textit{system-level} competence of a robot as it operates in OOD conditions. We highlight key research questions around this system-level view of OOD problems to guide future research toward safe and reliable learning-enabled autonomy.
\end{abstract}

\section{Introduction}\label{sec:introduction}
Machine learning (ML) systems are poised for widespread usage in robot autonomy stacks in the near future, driven by the successes of modern deep learning. 
For instance, decision-making algorithms in autonomous vehicles rely on ML-based perception and prediction models to estimate and forecast the state of the environment. 
As we increasingly rely on ML models to contend with the unstructured and unpredictable real world in robotics, it is paramount that we also acknowledge the shortcomings of our models, especially when we hope to deploy robots alongside humans in safety-critical settings.

In particular, ML models may behave unreliably on data that is dissimilar from the training data \textemdash{} inputs commonly termed \textit{out-of-distribution} (OOD).
This poses a significant challenge to deploying robots in the open world, e.g., as autonomous vehicles or home assistance robots, as such robots must interact with complex environments in conditions we cannot control or foresee. 
Coping with OOD inputs remains a key and largely unsolved challenge on the critical path to reliable and safe open-world autonomy. However, there is no generally-agreed-upon precise definition of what makes data OOD; instead, its definition is often left implicit and varies between problem formalisms and application contexts.

In this paper, we concretize the often nebulous notion of the OOD problem in robotics, drawing connections to existing approaches in the ML community. 
Critically, we advocate for a \textit{system-level} perspective of OOD data in robotics, which considers the impacts of OOD data on downstream decision making and leverages components throughout the full autonomy stack to mitigate negative consequences.
To this end, we present robotics research challenges at three timescales crucial to deploying reliable open-world autonomy: (i) real-time decision-making, (ii) episodic interaction with an environment, and (iii) the data lifecycle as learning-enabled robots are deployed, evaluated, and retrained. 

We note that this paper represents neither an algorithmic contribution nor a comprehensive survey of existing paradigms and literature on OOD topics in machine learning or robotics; in fact, many of the OOD topics that we discuss, like runtime-monitoring of perception systems \cite{RahmanCorkeEtAl2021} or heuristic uncertainty quantification of deep neural networks \cite{AbdarPourpanahEtAl2021}, constitute well-surveyed subfields in their own right. 
Rather than survey specific styles of analysis or approaches tailored towards particular submodules of the autonomy stack, our goal in this work is to provide an overview of the core considerations and system-wide challenges that we see as essential areas of robotics research activity for the coming years. 
Our contribution thus is to establish perspective and context to galvanize more research interest in a topic that we view as critical to improving the reliability of autonomous robots.
\begin{figure*}[]
\centering
\includegraphics[width=\textwidth]{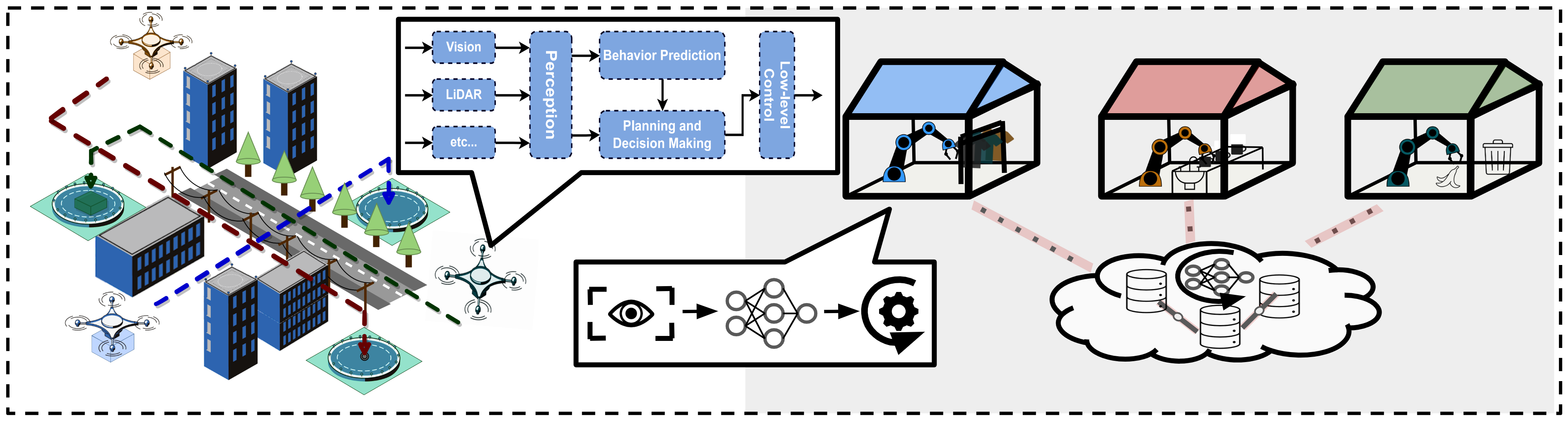}
\caption{\textbf{Left:} A future drone delivery system. It uses a modular autonomy stack consisting of 1) a perception system that builds an understanding of the drone's state and its environment, 2) a prediction module that makes inferences about the behavior of other agents and objects in the drone's environment, 3) a high-level planning and decision-making module, and 4) low-level controllers that actuate the drone's propellers to control the drone's trajectory. At deployment, the drone must safely navigate many obstacles such as other delivery drones, power lines, and trees, not all of which can be anticipated at design time. \textbf{Right:} Robotic manipulators assisting in the home. The manipulators are controlled end-to-end by directly passing the observations of the robot through a policy network that outputs the actions the robot should take. At deployment, the manipulators perform a wide range of household tasks like folding and hanging clothes, washing dishes, and cleaning up trash. Therefore, engineers train the manipulator policy by creating targeted experiments to collect a large and diverse training dataset of many objects to manipulate and tasks to complete. 
}
\label{fig:examples}
\end{figure*}

\section{Running Examples}
To better describe the challenges that OOD data creates in learning-enabled robotic systems, we use the two future autonomy systems shown in Figure \ref{fig:examples} as running examples in this paper. These conceptual examples highlight the plurality of applications and design paradigms used to leverage ML in the design of robotic systems. 

\textbf{Autonomous Drone Delivery Service:} Firstly, we consider an autonomous drone delivering packages in a city. As illustrated in Figure \ref{fig:examples}, this robot uses several learning-enabled components in its autonomy stack. The delivery drone has to make explainable decisions and meet stringent safety requirements by regulatory agencies to be deployed among humans. Crucially, to maintain these reliability requirements in rare and unforeseen circumstances, the drone needs mechanisms to detect and manage OOD inputs.

\textbf{Robotic Manipulators Assisting in the Home:} Secondly, we consider the deployment of robotic manipulators to assist with various tasks in and around the home, as shown in Figure \ref{fig:examples}. The manipulators' tasks are so diverse and unstructured that we consider a general manipulation policy trained in an end-to-end fashion in a controlled environment, as commonly considered in the reinforcement learning (RL) community. When we deploy these manipulators in people's homes, the environments and contexts that these robots encounter invariably differ from the lab or simulated environments used for training, which can markedly impact the system's performance. Therefore, ensuring reliable performance in OOD test environments is a crucial aspect of the design challenge. 

\section{What Makes Data Out-Of-Distribution?}

\begin{figure*}[t]
\centering
\includegraphics[width=\textwidth]{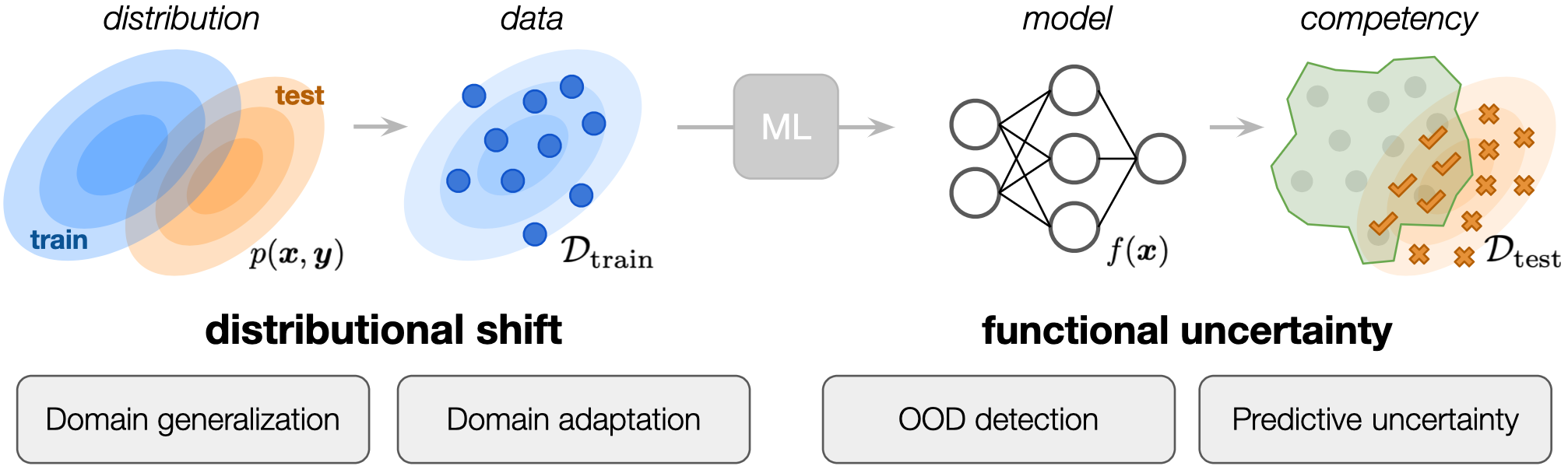}
\caption{Learning a predictive model $\learnedmodel$ from a finite dataset poses challenges, especially in the presence of distributional shift. To address this, methods in the ML community consider both training and adapting models in anticipation of and response to distributional shift. Besides improving models on OOD data, other approaches consider methods that quantify functional uncertainty by predicting when inputs are anomalous or quantifying uncertainty in the model's predictions.
\vspace{-0.2em}}

\label{fig:ml-methods}
\end{figure*}
Well-engineered ML pipelines produce models that generalize well to test data sampled i.i.d.\ from the same distribution as the training data. Consequently, when models \textit{fail} to generalize at test time, we often attribute this to 
``OOD data'' in a catch-all manner. 
What makes data OOD, and what causes these failures? 
In this section, we illustrate two concepts that structure our perspective on these questions using the notation of a standard supervised learning pipeline. Assume access to independent samples $\traindata = \{ (\x_i, \y_i) \}_{i=1}^\ndata$ drawn from an underlying joint distribution $\Ptrain$ with density $\ptrain(\x, \y)$, where $\x \in X$ and $\y \in Y$. In supervised learning, we fit a model $\learnedmodel: X \to Y$ on $\traindata$ and evaluate its performance on a test data set $\testdata$ drawn from $\Ptest$ with density $\ptest(\x, \y)$.

\textbf{Distributional Shifts:}
A \emph{distributional shift} occurs when test data $\testdata$ is sampled from a distribution $\Ptest$ that differs from $\Ptrain$, thereby making $\testdata$ OOD and $\traindata$ \emph{in-distribution} (ID). Shifts can corrupt the performance of the learned model $\learnedmodel$ since it may no longer capture the relationship between $\x$ and $\y$ in the test data. 
Distributional shifts can reflect fundamental changes in the underlying data generating process (often termed \emph{concept shift}). Concepts can shift discontinuously, like when important unobserved features change between train and test, or they can slowly drift over time. For example, when we use a model-based approach for lower-level control of the delivery drone, slowly degrading actuators can make the predictions of a learned dynamics model dangerously inaccurate.
Alternatively, shifts can be limited to part of the generative process. A \textit{covariate shift} describes when $p(\x)$ changes while $p(\y \mid \x)$ remains constant \cite{Shimodaira2000}. For instance, we might train a vision model for the delivery drone on images collected during the day but deploy the delivery drone using the model at night. Similarly, \textit{label shift} occurs when $p(\y)$ changes and $p(\x \mid \y)$ does not, for example when deploying a pre-trained pedestrian detector in a new country where there are overall more pedestrians \cite{SaerensLatinneEtAl2002}. 
The language of distributional shifts is particularly suited to quantifying how population level statistics, like the expected loss of a model, are affected by changing conditions.

\textbf{Functional Uncertainty:}
Since we do not have access to $\Ptest$ directly and must learn a model $\learnedmodel$ from a finite set of samples $\traindata$, we cannot be certain that $\learnedmodel$ will make good predictions at test time.
This offers a complementary view on the OOD problem; instead of reasoning about distributional differences,
we aim to characterize the domain of competence of a particular $\learnedmodel$, i.e., when and where we can have confidence in its individual predictions, and conversely, when we are uncertain in its predictions. We refer to this as the \textit{functional uncertainty} perspective on the OOD problem. 
Causes of high functional uncertainty are not rooted only in distributional notions; even when $\Ptest = \Ptrain$, the model $\learnedmodel$ may make poor predictions on rare inputs which were not well represented in the finite $\traindata$ \cite{GueringDelmasEtAl2023}. 
Instead, functional uncertainty may arise from \textit{epistemic} uncertainty, i.e., when we are unaware of the input-output relations that our models do not capture in the test domain.
Of course, distributional shifts can increase the likelihood of encountering test data outside the domain of competence of $\learnedmodel$.
Importantly, functional uncertainty is linked to how $\learnedmodel$ is used, as evaluating competence requires a measure of test-time performance. While this measure can be generic (e.g., KL divergence between the softmax output probabilities of  $\learnedmodel(\x)$ and $\ptest(\y \mid \x)$), it can also be tailored to downstream utility functions like the drone's landing performance.

\section{Trends in OOD in Machine Learning}\label{sec:oodinml}
The OOD problem is an open challenge in the ML community. Indeed, state-of-the-art models have been shown to be extremely sensitive to subtle distributional shifts (e.g., see  \cite{TorralbaEfros2011, GeirhosJacobsenEtAl2020, HendrycksDietterich2019, RechtRoelofsEtAl2019, MillerTaoriEtAl2021} and the references therein).
In this section, we discuss classical and core formulations and techniques from the ML literature that guide the ML community's approach to tackling the OOD challenge, summarized in Figure \ref{fig:ml-methods}.

\subsection{Coping with Distributional Shift}
Standard ML techniques are built around the often unrealistic assumption that $\Ptest = \Ptrain$. A major line of ML research aims to relax this assumption to develop learning algorithms that can cope with distributional shifts.

\textbf{Domain generalization} 
considers the capacity of a model trained only on data from $\Ptrain$ to generalize to an \emph{unknown} test-time data distribution $\Ptest$.
Thus, domain generalization amounts to coping with distributional shift between train and test time. 
To make this problem approachable, we need to make assumptions on how much $\Ptest$ can reasonably differ from $\Ptrain$. For example, one salient research direction aims to improve \textit{distributional robustness}, optimizing the worst-case performance within an envelope of distributional shifts to guarantee OOD performance \cite{Ben-TalHertogEtAl2013, DuchiNamkoong2021}. However, it is often unclear how large distributional uncertainty sets should be when conditions shift, a topic that roboticists working on applications should address. To circumvent such ambiguity, it is common to consider the robustness behavior on subpopulations of the training data instead \cite{SagawaKohEtAl2020}. A complementary research direction targets the root cause of poor generalization under distributional shift from a \textit{causal inference} perspective: Learned models often pick up spurious correlations in $\traindata$, rather than the invariant cause and effect relations that govern the underlying process \cite{Pearl2009,PetersBuhlmann2016,ArjovskyBottouEtAl2019}. For example, the vision-based robotic manipulator could rely on features in the background of the image to identify object types in the foreground (e.g, cooking items usually appear in kitchens), and thus fail to generalize to reasonable distribution shifts where the background changes (e.g., a pan on a sofa). 
Empirically, domain generalization often improves most when we (pre)train larger models on larger, more diverse datasets and infuse domain knowledge by encoding invariances explicitly or via data augmentations and  self-supervised pretraining \cite{MillerTaoriEtAl2021, ZhouLiu2022, GulrajaniLopes2021}. 
In light of these trends, leveraging foundation models (FMs) like large language models (LLMs) to incorporate expansive, generalist external knowledge presents a promising strategy for improving domain generalization \cite{BommasaniEtAl2021}. However, bridging the gap between an FM's training data---like large text corpora for LLMs---and a robot's observations (e.g., LiDAR) remains challenging \cite{BrohanBrownEtAl2023}. 

\textbf{Domain adaptation}  
aims to develop algorithms that leverage both the training dataset and some (usually unlabeled) test inputs \smash{$\{ \x_i \}_{i=1}^M \iid \Ptest$} to optimize the learned model $\learnedmodel$ on $\Ptest$. Domain adaptation therefore typically requires a priori availability of some test domain data. This can occur when we make batched predictions on a given test set, or, for example, when we deploy the drone delivery service in a new country and have a small budget for running pre-deployment trials.
Adapting $\learnedmodel$ to the test inputs is a paradigm that often yields drastic performance improvements with simple algorithms because the test domain data allows us to make inferences about a model's performance on $\Ptest$.  
For example, for covariate shift problems, the most elementary approach is to apply importance reweighting techniques to yield unbiased estimates of the model's risk under $\Ptest$ \cite{Shimodaira2000}. 
Classic results in domain adaptation theory state that performance degradation under distributional shift is linked to how well a classifier can distinguish data from the train and test domains \cite{Ben-DavidBlitzeretal2010, RedkoMorvantEtAl2020}.
These results motivate methods which learn feature representations such that train and test data look similar \cite{GaninUstinovaEtAl2016}, or learn transformations between the train and test domains \cite{HoffmanTzengEtAl2018}. 
More broadly, progress on algorithms that adapt models in response to shifted conditions is not limited to the batched setting (a setting that \cite{WilsonCook2020} surveys extensively). In particular, \emph{continual}, or \emph{lifelong}, learning algorithms seek to adapt $\learnedmodel$ over time in response to evolving distributions \cite{LangeAljundi2022}, and \emph{meta-learning} considers the design of algorithms that can rapidly adapt models to new distributions given separate datasets from related domains \cite{FinnAbeel2017}.
\subsection{Assessing Functional Uncertainty}
Domain adaptation and generalization focus on methods to select or improve the learned model $\learnedmodel$ in anticipation of or response to a changed data distribution $\Ptest$.
Orthogonally, we can also consider methods that aim to characterize the functional uncertainty of a \textit{particular} model $\learnedmodel$ trained on $\traindata$.

\textbf{Detecting anomalous inputs:}
A key source of functional uncertainty lies in inputs that are dissimilar to those seen in the training data. \textit{Anomaly detection} considers the challenge of predicting if an individual input is dissimilar to~$\traindata$ \cite{SalehiMirzaeiEtAl2021,RuffKauffmanEtAl2021, YangZhouEtAl2021}. This problem is also often called \textit{out-of-distribution detection}, but many approaches do not explicitly model the training distribution, so we use the more generic term \emph{anomaly detection}. 
We can measure dissimilarity from the training data in various ways, such as via a distance metric, or by evaluating likelihoods under a learned parametric model of $\ptrain(\x)$. These strategies are often applied in a learned feature space instead of directly on the inputs because modeling distances and distributions can be difficult for high-dimensional inputs, such as images. 

\textbf{Predictive Uncertainty:} An alternative approach is to design a model $\learnedmodel$ that directly outputs a measure of confidence in its predictions.
To ensure confidence scores are correct, some approaches ensure a model's predictions are \textit{calibrated}, i.e. that the predictive uncertainty matches the model's error rate \cite{GuoPleissEtAl2017}. Others, like conformal inference, construct prediction intervals containing the correct label with high probability \cite{BalasubramanianHoEtAll}.
However, it is generally challenging to ensure that the confidence measures we use to assess functional uncertainty remain calibrated under distributional shift. For example, the softmax output distribution of classification networks is often confidently wrong on OOD data \cite{OvadiaFertigEtAl2019}. Therefore, many empirical studies investigate design choices that encourage high predicted uncertainty on inputs that are dissimilar to training inputs, such as specific model architectures, auxiliary losses, and regularizers \cite{AbdarPourpanahEtAl2021}. 
Besides calibration algorithms and design choices that encourage high predictive uncertainty on anomalous inputs, \textit{Bayesian ML} offers an appealing approach to assess functional uncertainty under covariate shifts. This is because Bayesian methods allow us to quantify \textit{epistemic uncertainty} 
by incorporating \emph{subjective} prior beliefs to yield a posterior distribution $p(\learnedmodel \mid \traindata)$ \cite{AbdarPourpanahEtAl2021}. However, scaling Bayesian methods to large models is a computationally challenging task. Therefore, many methods approximate the Bayesian posterior, for example, through ensembling, or monte-carlo dropout \cite{LakshminarayananPritzelEtAll2017, GalZoubin2016}.

\subsection{Evaluation}
Researchers have developed benchmark datasets that contain train/test splits curated for qualitative semantic differences for evaluating OOD performance (e.g., see \cite{HendrycksDietterich2019,KohSagawaEtAl2021,MillerTaoriEtAl2021}). OOD test sets can include synthetic corruptions like motion blur, Gaussian noise, and other perturbations. In addition, many datasets test robustness to naturally occurring distribution shifts, like when we train the delivery drone's bird detection model only on images of land birds and include images of waterbirds in the test set, with some emphasizing tasks relevant to robotics (e.g., see \cite{SunSegoEtAl2022} and the references therein). 
Such data sets provide an intuitive foothold to develop algorithms by isolating reliability problems rooted in OOD data. Furthermore, some works introduce comparison suites for recent state-of-the-art algorithms to facilitate benchmarking on domain generalization \cite{GulrajaniLopes2021} or OOD detection \cite{ZhangYangEtAl2023} on simple tasks like image classification. 
However, it is often unclear how methods tested on semantically OOD datasets will impact a robotic system downstream at deployment.

\section{Open Challenges for OOD in Robotics}\label{sec:robotdifferent}

\begin{figure*}[t]
\centering
\includegraphics[width=\textwidth]{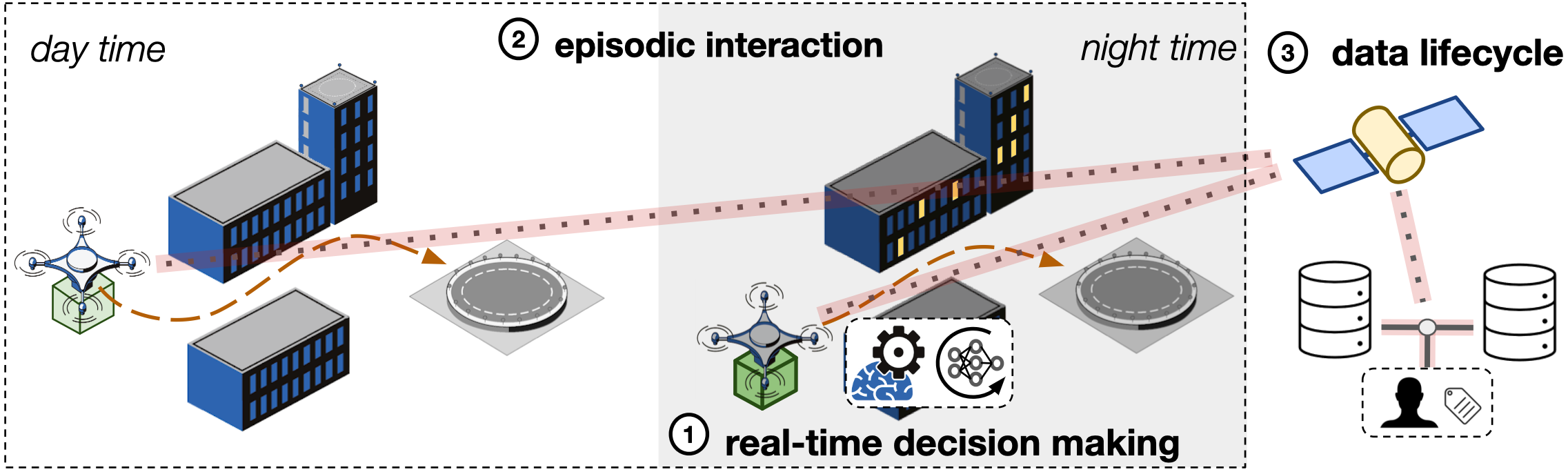}
\caption{Data-driven systems operating at different timescales. 
(1) Learning-enabled robotic systems must take actions and react to novel conditions in changing environments, requiring real-time OOD monitoring tools. 
(2) Long-horizon tasks (e.g.,  transporting a payload to a destination) require robotics OOD tools that consider episodic interactions, in which the typical assumption that inputs are drawn i.i.d. does not hold and time correlations should be accounted for. (3) Finally, learning-based models should be retrained offline to continuously improve the reliability of the overall robotic stack.}
\label{fig:timescales}
\end{figure*}

Robotics has always been centered on building \textit{systems} that work well in the real world. Therefore, we argue for a \textit{system-level} perspective on tackling OOD data in learning-enabled autonomy: Our ultimate goal is to reason about an ML-enabled autonomous system's reliability and competence when it applies learned models in a feedback loop over time, as it operates in potentially shifted conditions. This perspective differs from the model-level paradigms in the ML community aimed at quantifying how a model's accuracy degrades on independently-sampled OOD data because learned models only constitute individual components of a complex autonomy stack. Therefore, system-level perspectives present unique challenges for the robotics community related to 1) detecting OOD conditions, 2) responding to them to avert system failures, and 3) improving the robotic system's OOD closed-loop performance as a whole.
We illustrate these challenges by considering three different timescales at which data-driven robotic systems operate, as shown in Figure \ref{fig:timescales}, each with distinct OOD challenges for robotics. We discuss each timescale, drawing connections to methods from the ML community and highlighting key open research questions (\RQ{}s) toward autonomous systems that leverage ML while being robust to the OOD conditions they will inevitably encounter. In addition, we examine various aspects of the RQs using our running examples and briefly touch upon recent research trends to contextualize the RQs. We emphasize that this discussion is not an exhaustive survey of existing literature but rather a brief discussion to underscore the significance of the RQs.  

\subsection{Real-time ML-Enabled Decision Making}
To maintain system-level competence at runtime, we need to reason about the downstream impact of individual OOD inputs on the decision-making system in real-time. For example, a failure of the delivery drone to detect a pedestrian could be disastrous; we need to construct safeguards that ensure inference errors do not lead to system failure. Therefore, at the real-time timescale, we need to monitor the competence of the full decision-making stack on individual inputs encountered at test time. Even though runtime monitoring is commonplace in robotics, its application to mitigate the effect of OOD data on state-of-the-art learned components suggests two key research questions centered around the functional uncertainty viewpoint on OOD.

\begin{researchquestion}[Averting OOD failures through Runtime Monitoring]\label{rq:runtime}
Can we leverage \emph{full-stack} sensory information at runtime to detect if a decision system relying on a learned model $\learnedmodel$ will perform poorly, before a failure occurs?
\end{researchquestion}
Because we generally cannot identify all aspects of the robot's environment that affect the reliability of its ML components, provable safety guarantees are virtually unattainable for autonomous systems without making restrictive assumptions \cite{SeshiaSadighEtAl2016}. Indeed, the failures caused by conditions that were not represented at design time---be it in training data or in simulated test scenarios---are generally what we attribute as OOD. Therefore, we view algorithms for monitoring autonomous systems at runtime as a core aspect of maintaining system-level competence in OOD regimes. 

Assessing the functional uncertainty on the model's inputs is an important first step towards this goal, but is not sufficient to monitor the performance of the overall robotics stack.
Instead, we need to reason about how functional uncertainty propagates through the decision-making system and devise goal-oriented measures of uncertainty on $\learnedmodel$ that capture system-level performance. Indeed, the downstream impact of erroneous predictions may vary between systems or the current system state. 
Access to the full robotic autonomy stack also presents opportunities to use information from additional sensors besides the model's inputs to improve our assessment of functional uncertainty during operation.

\begin{researchquestion}[OOD Aware Decision Making]\label{rq:ood-dec}
Can we design decision-making systems compatible with runtime monitors robust to high functional uncertainty?
\end{researchquestion}
A robot must always choose an action to take, even if runtime monitors suggest that a learned component $\learnedmodel$ is operating OOD. Thus, as roboticists, we must design systems where model uncertainties are assessed and accounted for during decision-making. This entails the joint design of real-time runtime monitors, uncertainty-aware decision-making algorithms, and fallback strategies.
Since fallbacks may need to rely on redundancy or alternate sources of information, the problem of ensuring the safety and reliability of the aggregate autonomous system is a significantly more expansive challenge than that of characterizing the functional uncertainty of an ML model in isolation.

\textbf{Additional Examples:} Consider the scenario of the delivery drone's perception network failing to detect OOD object types not seen at training time. At high speeds, missing a detection on a single image can make obstacle avoidance impossible. Runtime monitors as suggested in \RQ{\ref{rq:runtime}} that flag when the functional uncertainty of the visual object detector on a specific input is high, signaling that the model outputs are unreliable, can be critical to avoid catastrophic failures. However, not every missed detection will affect the same consequences: a missed detection of an OOD bird breed far away is less likely to cause a collision than a missed detection of a nearby tree (complicating \RQ{\ref{rq:runtime}}). When the runtime monitor signals that the object detection system is dangerously inaccurate, the drone should land or continue safe operation in a degraded state. Certifying that a fallback strategy does not cause additional hazards, such as ensuring the drone does not land on busy roads with limited sensing, requires the system-level considerations outlined in \RQ{\ref{rq:ood-dec}}.

Knowing that particular objects are OOD for the robot manipulator can help it decide which objects it can reliably manipulate. This knowledge can allow the robot to abstain from handling OOD objects instead of dropping and damaging them. In addition, we can build system-level checks around the manipulator policy to sanity-check its decisions. For example, we could compare to grasps computed using more classical techniques or leverage additional sensing modalities to estimate when the robot can or cannot successfully manipulate an object.

\textbf{Recent Trends:} As discussed in Section \ref{sec:oodinml}, quantifying functional uncertainty of a model on OOD inputs is a lively field of study, including anomaly detection (as surveyed in \cite{SalehiMirzaeiEtAl2021, RuffKauffmanEtAl2021, HodgeJim2004}) and heuristics for predictive uncertainty like approximate Bayesian inference (e.g., \cite{SharmaAzizanEtAl2021, LakshminarayananPritzelEtAll2017, AminiSchwartingEtAl2020}) to name a few. 
However, such methods generally only apply to covariate shifts. In addition, robotics-focused monitoring methods that leverage additional sensors to learn how models are innacurate or check consistency among modules (as surveyed in \cite{RahmanCorkeEtAl2021}), or learn to predict or recognize system failures also show promise \cite{LuoZhaoEtAl2021, FaridSnyderEtAl2022}. However, these early approaches are often not goal-oriented, heuristic, or unverifiable under distribution shift. Moreover, while many existing control theoretic approaches provide safety filters that interfere with black-box learned policies to correct trajectories in settings where system dynamics and state are known (e.g., \cite{LeungSchmerlingEtAl2019, FisacAkametaluEtAl2019}), certifiably closing the loop on runtime monitors and fallbacks in complex systems like our drone delivery example is a broadly understudied problem.  

\subsection{Episodic closed-loop interaction}
Learning-enabled robots do not passively make predictions on a set of given individual inputs. Instead, they actively interact with their environment to perform tasks. Thus, reliable robotic systems should also reason about the influence of OOD conditions on the closed-loop decision-making system over extended periods of time. At this timescale, 
this \textit{sequential decision-making} context induces key distinct research challenges for the robotics community.
\begin{researchquestion}[Temporally Correlated OOD events]\label{rq:ep-cor}
Can we develop methods that account for the temporal correlations between inputs when we repeatedly evaluate a learned model $\learnedmodel$ under shifted conditions over the course of an episode?
\end{researchquestion}
As discussed in Section \ref{sec:oodinml}, considering population statistics like the expected loss under distributional shift is one of the core frameworks in ML research to study OOD performance.  
However, when we deploy a robot over an episode, the learned model's inputs will be correlated over time, violating the standard ML assumption that test samples are i.i.d. Even in nominal conditions, these temporal correlations induce distributional shifts from training data. For example, while an ML perception model would likely be trained on a set of shuffled inputs from diverse weather conditions from many trips, an autonomous vehicle will likely only encounter one weather condition during a particular trip. Therefore, as roboticists, we should investigate how we can strengthen performance in anticipation of shifted conditions that affect the reliability of model outputs over the course of a trajectory, for example, by assuring generalization across domains or rapidly adapting to conditions faced at deployment. In addition, we consider developing methods that certifiably detect performance-impacting shifts during execution without i.i.d. assumptions as a largely open problem.

\begin{researchquestion}[Mitigating Distributional Shifts]\label{rq:ep-dec}
Can we construct decision-making algorithms that mitigate distributional shifts between the training and deployment conditions to ensure the overall reliability of the deployed system?
\end{researchquestion}
Robotic systems often have agency to mitigate distributional shifts through decision-making.
For example, a drone can avoid aggressive maneuvers in regions where it has limited data to mitigate the consequences of potential errors in a learned dynamics model $f$. 
By ensuring that learning-enabled components operate in-distribution, the design of OOD monitors is simplified, the use of fallback strategies is reduced, and the reliability of the robotic stack is generally improved. To achieve this intelligent behavior, we must design methods to quantify and reason about the \textit{domain of competency} of learned systems in a manner that is amenable to planning and decision making.

\textbf{Additional Examples:} Externally shifting conditions, such as those that occur when we train the drone's vision system on daytime images and deploy at night, will consistently degrade the perception system. \RQ{\ref{rq:ep-cor}} asks how we can distinguish consistent model errors induced by OOD conditions from sporadic errors, which may be tolerable. How many contiguous missed detections will induce a failure, and how do we anticipate this before it is too late? In addition, distributional shifts can stem from the fact that the drone needs to use some policy for test deployments to collect data: The delivery drone might use a slow and conservative policy to collect the data to learn interaction models for other agents. If the drone flies very aggressively using these models, the closed-loop trajectory distribution will shift, making the interaction models dangerously inaccurate. Exploiting the drone's ability to control this shift is  \RQ{\ref{rq:ep-dec}}'s focus.

For the robotic manipulator, every household deployment represents shifted conditions from the environment in which we developed the policy: clothing styles and typical pots and pans vary between households and countries. In the context of \RQ{\ref{rq:ep-cor}} and \RQ{\ref{rq:ep-dec}}, we should study how we can quickly adapt the robot to the shifted or evolving conditions and expand operations to new task contexts reliably. 

\textbf{Recent Trends:} Uncertainty in a robot's dynamics model or surroundings are forms of temporally correlated OOD uncertainties eminent both in the learning-based control and RL communities through topics like dynamics learning and sim2real transfer \cite{BrunkeGreefEtAl2022}. However, quantifying and managing the system-level effects of temporally correlated OOD data in complex sensing modalities like perception systems is an expansive open problem, with early steps including contributions like \cite{FaridVeerEtAl2022, PodkopaevRamdasEtAl2022} aimed at detecting shifted distributions between task executions rather than during long-term deployments. In addition, the distributional shift induced by a change in data collection and test policies is a core framework through which many robot learning problems,
like imitation learning \cite{RossGordon2011} and
offline RL \cite{LevineKumarEtAl2020}, are studied. More broadly, ego-influenced distributional shifts may affect any learned model in an autonomy stack, not just systems trained end-to-end.

\subsection{Data lifecycle}
Finally, beyond interactions during individual episodes, it is also important to consider the long-term cycles over which data-driven robotic systems are deployed, evaluated, improved, and deployed again. In this context, we view the development process as a feedback loop, potentially with human experts in the loop. At this scale, our goal is to use data collected during operation to improve the system's overall performance across novel, rare, or shifted conditions. 

\begin{researchquestion}[Leveraging Operational Data]\label{rq:op-data}
How can we use data collected during operation in diverse tasks and contexts to improve the robustness and quality of learned models?
\end{researchquestion}
Retraining components on new data collected during operation can mitigate the influence of OOD conditions by reducing functional uncertainty and ensuring that training data matches test conditions. 
However, simply appending operational data to $\traindata$ may not be enough to avoid learning spurious correlations or improve performance on extremely rare failure modes.
Therefore, we should also aim to increase the diversity of the data and leverage the fact that data collected during different episodes of robot execution represents a set of diverse test-time contexts, which can be naturally grouped into different operational domains. This task-specific structure lends itself well to a variety of approaches to improve domain generalization, like distributionally robust, multi-task, meta-, or causal learning, which can yield a model that is 
able to better generalize to new conditions.

\begin{researchquestion}[Efficient Data Collection]\label{rq:data-col}
How do we select what operational data to use to efficiently improve our models?
\end{researchquestion}
Robotic fleets collect tremendous amounts of data during operation, not all of which can be stored or labeled to improve the performance of the autonomy stack. Moreover, collecting more data through robot deployments is costly and can see diminishing returns. Therefore, we need to understand how to efficiently collect data during operation or testing and judiciously choose which data to flag for labeling. This problem has strong connections to active learning and research on estimating functional uncertainty in ML models, as the most informative inputs to label correspond to those on which the model $f$ is most uncertain.

\textbf{Additional Examples:} 
Daily varying wind and weather conditions can significantly affect the dynamics of a delivery drone carrying a large payload in an a priori hard-to-model fashion. In line with \RQ{\ref{rq:op-data}}, we can use the episodically collected trajectory data more effectively by using meta-learning techniques to learn structure in the weather disturbance dynamics so we can adapt online more rapidly, improving control performance. However, a small delivery drone has severely limited data storage capacity. Therefore, \RQ{\ref{rq:data-col}} concerns how we should select which data to store to improve the system and what data we discard in real-time. Furthermore, it is insufficient to keep a buffer and upload it when an incident or failure occurs because reliability requirements make system failures extremely rare, even when learning-enabled subcomponents often make errors. 

Alternatively, consider an example where we train the robot manipulator to complete ten separate household tasks. Suppose we naively train to maximize the average performance across tasks. In that case, we might greedily sacrifice performance on one task, like sweeping the floor, because improvements on another task, like loading dishwashers, outweigh the cost. Then, during deployment, some users might request the robot to sweep the floor more frequently than they use it to load dishes. The robot will perform poorly on the resulting shifted task distribution, even though the engineers in the lab see a high average performance. This example illustrates that, in general, we may not succeed on rare or shifted conditions if we naively optimize performance according to prevalence in some dataset. Instead, as is the focus in (subgroup) distributional robustness or multi-task learning, we should train the manipulator in line with \RQ{\ref{rq:op-data}} so that performance is consistently good across tasks. Then, users will always observe good performance no matter their task proclivities. In addition, engineers will operate on a fixed budget for experimentation and training, so they must judiciously design experiments that maximize the robot's performance, a facet of \RQ{\ref{rq:data-col}}.

\textbf{Recent Trends:} Roboticists have already started leveraging some of the ML community trends to improve generalization when distributions vary across deployments beyond simply augmenting datasets with new operational data and retraining or, as surveyed in \cite{LesortLomonaco2020}, by applying techniques from continual learning: Some approaches learn consistent patterns across distributions by considering separate losses for each trajectory, for example, by applying meta-learning to more rapidly learn dynamics models \cite{RichardsAzizanEtAl2021} or policies \cite{NagabandiClaveraEtAl2019} across environments, leveraging causal inference techniques to identify generalizable state and task representations (as surveyed in \cite{StokingGopnikEtAl2022}, \cite{KirkZhangEtAl2021}), or through multi-task learning \cite{RusuColmenarejoEtAl2016}.
Finally, active learning techniques \cite{Settles2012} need to be tailored to robotics to collect and label data efficiently. Ongoing efforts in these areas underscore the significance of the data lifecycle challenges in robotics and make progress on specific components in the autonomy stack. 

\subsection{Evaluation}
As the ML community showed, establishing accessible simulation platforms to benchmark closed-loop OOD performance will accelerate progress. Given the breadth of the RQs, it is clear that we require a diverse suite of testbeds, each targeting specific aspects of the RQs. A persistent challenge is the limited flexibility of common simulators and the effort required to develop nominally reliable robots, which often restricts experimentation to rudimentary interventions on a robot's environment, like exposing a highway AV planner to roundabouts \cite{FilosTigas2020}, changing lighting conditions from day to night \cite{SharmaAzizanEtAl2021}, swapping cups with bowls \cite{FaridVeerEtAl2022}, or one-off hardware experiments like navigating a rover through novel hallways \cite{RichterRoy2017}. Therefore, an essential task is developing community benchmarks with held-out OOD scenarios that more closely reflect the organic failure modes a mature system may experience throughout varied real-world deployments.

\section{Conclusion}\label{sec:conclusion}
The recurring theme across these timescales is that the \textit{full-stack} nature of robotics requires a \textit{system-level} perspective on the OOD problem. We argue that roboticists should embrace this system-level perspective: Investigate both how OOD data impacts the reliability of the full autonomy stack and how to leverage the full autonomy stack to mitigate negative consequences. 
While these research questions are challenging and involve all aspects of the autonomy stack, they represent necessary steps towards a future where we can safely and reliably leverage ML to enable true open-world autonomy.

\section*{Acknowledgements}
The NASA University Leadership initiative (grant \#80NSSC20M0163) and KACST provided funds to assist the authors with their research, but this article solely reflects the opinions and conclusions of its authors and not any NASA or KACST entity.

\bibliographystyle{unsrt}
\bibliography{references}

\begin{thebibliography}{10}

\bibitem{RahmanCorkeEtAl2021}
Quazi~Marufur Rahman, Peter Corke, and Feras Dayoub.
\newblock Run-time monitoring of machine learning for robotic perception: A
  survey of emerging trends.
\newblock {\em IEEE Access}, 9:20067--20075, 2021.

\bibitem{AbdarPourpanahEtAl2021}
Moloud Abdar, Farhad Pourpanah, Sadiq Hussain, Dana Rezazadegan, Li~Liu,
  Mohammad Ghavamzadeh, Paul Fieguth, Xiaochun Cao, Abbas Khosravi, U.~Rajendra
  Acharya, Vladimir Makarenkov, and Saeid Nahavandi.
\newblock A review of uncertainty quantification in deep learning: Techniques,
  applications and challenges.
\newblock {\em Information Fusion}, 76:243--297, 2021.

\bibitem{Shimodaira2000}
H.~Shimodaira.
\newblock Improving predictive inference under covariate shift by weighting the
  log-likelihood function.
\newblock {\em Journal of Statistical Planning and Inference}, 90:227--244,
  2000.

\bibitem{SaerensLatinneEtAl2002}
Marco Saerens, Patrice Latinne, and Christine Decaestecker.
\newblock Adjusting the outputs of a classifier to new a priori probabilities:
  A simple procedure.
\newblock {\em Neural Computation}, 14:21--41, 2002.

\bibitem{GueringDelmasEtAl2023}
Joris Guerin, Kevin Delmas, Raul Ferreira, and Jérémie Guiochet.
\newblock Out-of-distribution detection is not all you need.
\newblock {\em Proceedings of the AAAI Conference on Artificial Intelligence},
  37(12):14829--14837, Jun. 2023.

\bibitem{TorralbaEfros2011}
Antonio Torralba and Alexei~A. Efros.
\newblock Unbiased look at dataset bias.
\newblock In {\em CVPR 2011}, pages 1521--1528, 2011.

\bibitem{GeirhosJacobsenEtAl2020}
Robert Geirhos, J{\"o}rn-Henrik Jacobsen, Claudio Michaelis, Richard Zemel,
  Wieland Brendel, Matthias Bethge, and Felix~A. Wichmann.
\newblock Shortcut learning in deep neural networks.
\newblock {\em Nature Machine Intelligence}, 2(11):665--673, Nov 2020.

\bibitem{HendrycksDietterich2019}
Dan Hendrycks and Thomas Dietterich.
\newblock Benchmarking neural network robustness to common corruptions and
  perturbations.
\newblock {\em Proceedings of the International Conference on Learning
  Representations}, 2019.

\bibitem{RechtRoelofsEtAl2019}
Benjamin Recht, Rebecca Roelofs, Ludwig Schmidt, and Vaishaal Shankar.
\newblock Do {I}mage{N}et classifiers generalize to {I}mage{N}et?
\newblock In Kamalika Chaudhuri and Ruslan Salakhutdinov, editors, {\em
  Proceedings of the 36th International Conference on Machine Learning},
  volume~97 of {\em Proceedings of Machine Learning Research}, pages
  5389--5400. PMLR, 09--15 Jun 2019.

\bibitem{MillerTaoriEtAl2021}
John~P Miller, Rohan Taori, Aditi Raghunathan, Shiori Sagawa, Pang~Wei Koh,
  Vaishaal Shankar, Percy Liang, Yair Carmon, and Ludwig Schmidt.
\newblock Accuracy on the line: on the strong correlation between
  out-of-distribution and in-distribution generalization.
\newblock In Marina Meila and Tong Zhang, editors, {\em Proceedings of the 38th
  International Conference on Machine Learning}, volume 139 of {\em Proceedings
  of Machine Learning Research}, pages 7721--7735. PMLR, 18--24 Jul 2021.

\bibitem{Ben-TalHertogEtAl2013}
Aharon Ben-Tal, Dick den Hertog, Anja~De Waegenaere, Bertrand Melenberg, and
  Gijs Rennen.
\newblock Robust solutions of optimization problems affected by uncertain
  probabilities.
\newblock {\em Management Science}, 59(2):341--357, 2013.

\bibitem{DuchiNamkoong2021}
John~C. Duchi and Hongseok Namkoong.
\newblock {Learning models with uniform performance via distributionally robust
  optimization}.
\newblock {\em The Annals of Statistics}, 49(3):1378 -- 1406, 2021.

\bibitem{SagawaKohEtAl2020}
Shiori Sagawa, Pang~Wei Koh*, Tatsunori~B. Hashimoto, and Percy Liang.
\newblock Distributionally robust neural networks.
\newblock In {\em International Conference on Learning Representations}, 2020.

\bibitem{Pearl2009}
Judea Pearl.
\newblock {\em Causality}.
\newblock Cambridge University Press, 2009.

\bibitem{PetersBuhlmann2016}
Jonas Peters, Peter Bühlmann, and Nicolai Meinshausen.
\newblock Causal inference by using invariant prediction: identification and
  confidence intervals.
\newblock {\em Journal of the Royal Statistical Society: Series B (Statistical
  Methodology)}, 78(5):947--1012, 2016.

\bibitem{ArjovskyBottouEtAl2019}
Martin Arjovsky, Léon Bottou, Ishaan Gulrajani, and David Lopez-Paz.
\newblock Invariant risk minimization.
\newblock {\em arxiv preprint arxiv:1907.02893}, 2020.

\bibitem{ZhouLiu2022}
Kaiyang Zhou, Ziwei Liu, Yu~Qiao, Tao Xiang, and Chen~Change Loy.
\newblock Domain generalization: A survey.
\newblock {\em IEEE Transactions on Pattern Analysis and Machine Intelligence},
  pages 1--20, 2022.

\bibitem{GulrajaniLopes2021}
Ishaan Gulrajani and David Lopez-Paz.
\newblock In search of lost domain generalization.
\newblock In {\em International Conference on Learning Representations}, 2021.

\bibitem{BommasaniEtAl2021}
Rishi Bommasani, Drew~A. Hudson, Ehsan Adeli, Russ Altman, Simran Arora, Sydney
  von Arx, Michael~S. Bernstein, Jeannette Bohg, Antoine Bosselut, Emma
  Brunskill, Erik Brynjolfsson, S.~Buch, Dallas Card, Rodrigo Castellon,
  Niladri~S. Chatterji, Annie~S. Chen, Kathleen~A. Creel, Jared Davis, Dora
  Demszky, Chris Donahue, Moussa Doumbouya, Esin Durmus, Stefano Ermon, John
  Etchemendy, Kawin Ethayarajh, Li~Fei-Fei, Chelsea Finn, Trevor Gale,
  Lauren~E. Gillespie, Karan Goel, Noah~D. Goodman, Shelby Grossman, Neel Guha,
  Tatsunori Hashimoto, Peter Henderson, John Hewitt, Daniel~E. Ho, Jenny Hong,
  Kyle Hsu, Jing Huang, Thomas~F. Icard, Saahil Jain, Dan Jurafsky, Pratyusha
  Kalluri, Siddharth Karamcheti, Geoff Keeling, Fereshte Khani, O.~Khattab,
  Pang~Wei Koh, Mark~S. Krass, Ranjay Krishna, Rohith Kuditipudi, Ananya Kumar,
  Faisal Ladhak, Mina Lee, Tony Lee, Jure Leskovec, Isabelle Levent, Xiang~Lisa
  Li, Xuechen Li, Tengyu Ma, Ali Malik, Christopher~D. Manning, Suvir~P.
  Mirchandani, Eric Mitchell, Zanele Munyikwa, Suraj Nair, Avanika Narayan,
  Deepak Narayanan, Benjamin Newman, Allen Nie, Juan~Carlos Niebles, Hamed
  Nilforoshan, J.~F. Nyarko, Giray Ogut, Laurel Orr, Isabel Papadimitriou,
  Joon~Sung Park, Chris Piech, Eva Portelance, Christopher Potts, Aditi
  Raghunathan, Robert Reich, Hongyu Ren, Frieda Rong, Yusuf~H. Roohani, Camilo
  Ruiz, Jack Ryan, Christopher R'e, Dorsa Sadigh, Shiori Sagawa, Keshav
  Santhanam, Andy Shih, Krishna~Parasuram Srinivasan, Alex Tamkin, Rohan Taori,
  Armin~W. Thomas, Florian Tram{\`e}r, Rose~E. Wang, William Wang, Bohan Wu,
  Jiajun Wu, Yuhuai Wu, Sang~Michael Xie, Michihiro Yasunaga, Jiaxuan You,
  Matei~A. Zaharia, Michael Zhang, Tianyi Zhang, Xikun Zhang, Yuhui Zhang,
  Lucia Zheng, Kaitlyn Zhou, and Percy Liang.
\newblock On the opportunities and risks of foundation models.
\newblock {\em ArXiv}, 2021.

\bibitem{BrohanBrownEtAl2023}
Anthony Brohan, Noah Brown, Justice Carbajal, Yevgen Chebotar, Xi~Chen,
  Krzysztof Choromanski, Tianli Ding, Danny Driess, Avinava Dubey, Chelsea
  Finn, Pete Florence, Chuyuan Fu, Montse~Gonzalez Arenas, Keerthana
  Gopalakrishnan, Kehang Han, Karol Hausman, Alexander Herzog, Jasmine Hsu,
  Brian Ichter, Alex Irpan, Nikhil Joshi, Ryan Julian, Dmitry Kalashnikov,
  Yuheng Kuang, Isabel Leal, Lisa Lee, Tsang-Wei~Edward Lee, Sergey Levine, Yao
  Lu, Henryk Michalewski, Igor Mordatch, Karl Pertsch, Kanishka Rao, Krista
  Reymann, Michael Ryoo, Grecia Salazar, Pannag Sanketi, Pierre Sermanet,
  Jaspiar Singh, Anikait Singh, Radu Soricut, Huong Tran, Vincent Vanhoucke,
  Quan Vuong, Ayzaan Wahid, Stefan Welker, Paul Wohlhart, Jialin Wu, Fei Xia,
  Ted Xiao, Peng Xu, Sichun Xu, Tianhe Yu, and Brianna Zitkovich.
\newblock Rt-2: Vision-language-action models transfer web knowledge to robotic
  control.
\newblock {\em arXiv preprint arXiv:2307.15818}, 2023.

\bibitem{Ben-DavidBlitzeretal2010}
S.~Ben-David, J.~Blitzer, K.~Crammer, Alex Kulesza, Fernando Pereira, and
  Jennifer Wortman~Vaughan.
\newblock {\em A theory of learning from different domains}.
\newblock Machine Learning 79, 2010.

\bibitem{RedkoMorvantEtAl2020}
Ievgen Redko, Emilie Morvant, Amaury Habrard, Marc Sebban, and Younès Bennani.
\newblock A survey on domain adaptation theory.
\newblock {\em arxiv preprint, arxiv:2004.11829}, 2020.

\bibitem{GaninUstinovaEtAl2016}
Yaroslav Ganin, Evgeniya Ustinova, Hana Ajakan, Pascal Germain, Hugo
  Larochelle, Fran\c{c}ois Laviolette, Mario Marchand, and Victor Lempitsky.
\newblock Domain-adversarial training of neural networks.
\newblock {\em J. Mach. Learn. Res.}, 17(1):2096–2030, jan 2016.

\bibitem{HoffmanTzengEtAl2018}
Judy Hoffman, Eric Tzeng, Taesung Park, Jun-Yan Zhu, Phillip Isola, Kate
  Saenko, Alexei Efros, and Trevor Darrell.
\newblock {C}y{CADA}: Cycle-consistent adversarial domain adaptation.
\newblock In {\em Proceedings of the 35th International Conference on Machine
  Learning}, volume~80 of {\em Proceedings of Machine Learning Research}, pages
  1989--1998. PMLR, 10--15 Jul 2018.

\bibitem{WilsonCook2020}
Garrett Wilson and Diane~J. Cook.
\newblock A survey of unsupervised deep domain adaptation.
\newblock {\em ACM Trans. Intell. Syst. Technol.}, 11(5), jul 2020.

\bibitem{LangeAljundi2022}
Matthias De~Lange, Rahaf Aljundi, Marc Masana, Sarah Parisot, Xu~Jia, Aleš
  Leonardis, Gregory Slabaugh, and Tinne Tuytelaars.
\newblock A continual learning survey: Defying forgetting in classification
  tasks.
\newblock {\em IEEE Transactions on Pattern Analysis and Machine Intelligence},
  44(7):3366--3385, 2022.

\bibitem{FinnAbeel2017}
Chelsea Finn, Pieter Abbeel, and Sergey Levine.
\newblock Model-agnostic meta-learning for fast adaptation of deep networks.
\newblock In {\em Proceedings of the 34th International Conference on Machine
  Learning}, volume~70 of {\em Proceedings of Machine Learning Research}, pages
  1126--1135. PMLR, 06--11 Aug 2017.

\bibitem{SalehiMirzaeiEtAl2021}
Mohammadreza Salehi, Hossein Mirzaei, Dan Hendrycks, Yixuan Li,
  Mohammad~Hossein Rohban, and Mohammad Sabokrou.
\newblock A unified survey on anomaly, novelty, open-set, and
  out-of-distribution detection: Solutions and future challenges.
\newblock {\em arxiv preprint arxiv:2110.14051}, 2021.

\bibitem{RuffKauffmanEtAl2021}
Lukas Ruff, Jacob~R. Kauffmann, Robert~A. Vandermeulen, Grégoire Montavon,
  Wojciech Samek, Marius Kloft, Thomas~G. Dietterich, and Klaus-Robert Müller.
\newblock A unifying review of deep and shallow anomaly detection.
\newblock {\em Proceedings of the IEEE}, 109(5):756--795, 2021.

\bibitem{YangZhouEtAl2021}
Jingkang Yang, Kaiyang Zhou, Yixuan Li, and Ziwei Liu.
\newblock Generalized out-of-distribution detection: A survey.
\newblock {\em arXiv preprint arXiv:2110.11334}, 2021.

\bibitem{GuoPleissEtAl2017}
Chuan Guo, Geoff Pleiss, Yu~Sun, and Kilian~Q. Weinberger.
\newblock On calibration of modern neural networks.
\newblock In {\em Proceedings of the 34th International Conference on Machine
  Learning - Volume 70}, ICML'17, page 1321–1330. JMLR.org, 2017.

\bibitem{BalasubramanianHoEtAll}
Vineeth~N. Balasubramanian, Shen-Shyang Ho, and Vladimir Vovk.
\newblock {\em Conformal Prediction for Reliable Machine Learning}.
\newblock Morgan Kaufmann, 2014.

\bibitem{OvadiaFertigEtAl2019}
Yaniv Ovadia, Emily Fertig, Jie Ren, Zachary Nado, D.~Sculley, Sebastian
  Nowozin, Joshua~V. Dillon, Balaji Lakshminarayanan, and Jasper Snoek.
\newblock Can you trust your model's uncertainty? evaluating predictive
  uncertainty under dataset shift.
\newblock In {\em Proceedings of the 33rd International Conference on Neural
  Information Processing Systems}, Red Hook, NY, USA, 2019. Curran Associates
  Inc.

\bibitem{LakshminarayananPritzelEtAll2017}
Balaji Lakshminarayanan, Alexander Pritzel, and Charles Blundell.
\newblock Simple and scalable predictive uncertainty estimation using deep
  ensembles.
\newblock In I.~Guyon, U.~Von Luxburg, S.~Bengio, H.~Wallach, R.~Fergus,
  S.~Vishwanathan, and R.~Garnett, editors, {\em Advances in Neural Information
  Processing Systems}, volume~30. Curran Associates, Inc., 2017.

\bibitem{GalZoubin2016}
Yarin Gal and Zoubin Ghahramani.
\newblock Dropout as a bayesian approximation: Representing model uncertainty
  in deep learning.
\newblock In Maria~Florina Balcan and Kilian~Q. Weinberger, editors, {\em
  Proceedings of The 33rd International Conference on Machine Learning},
  volume~48 of {\em Proceedings of Machine Learning Research}, pages
  1050--1059, New York, New York, USA, 20--22 Jun 2016. PMLR.

\bibitem{KohSagawaEtAl2021}
Pang~Wei Koh et~al.
\newblock Wilds: A benchmark of in-the-wild distribution shifts.
\newblock In Marina Meila and Tong Zhang, editors, {\em Proceedings of the 38th
  International Conference on Machine Learning}, volume 139 of {\em Proceedings
  of Machine Learning Research}, pages 5637--5664. PMLR, 18--24 Jul 2021.

\bibitem{SunSegoEtAl2022}
Tao Sun, Mattia Segu, Janis Postels, Yuxuan Wang, Luc Van~Gool, Bernt Schiele,
  Federico Tombari, and Fisher Yu.
\newblock Shift: A synthetic driving dataset for continuous multi-task domain
  adaptation.
\newblock In {\em Proceedings of the IEEE/CVF Conference on Computer Vision and
  Pattern Recognition (CVPR)}, pages 21371--21382, June 2022.

\bibitem{ZhangYangEtAl2023}
Jingyang Zhang, Jingkang Yang, Pengyun Wang, Haoqi Wang, Yueqian Lin, Haoran
  Zhang, Yiyou Sun, Xuefeng Du, Kaiyang Zhou, Wayne Zhang, Yixuan Li, Ziwei
  Liu, Yiran Chen, and Hai Li.
\newblock Openood v1.5: Enhanced benchmark for out-of-distribution detection.
\newblock {\em arXiv preprint arXiv:2306.09301}, 2023.

\bibitem{SeshiaSadighEtAl2016}
Sanjit~A. Seshia, Dorsa Sadigh, and S.~Shankar Sastry.
\newblock Towards verified artificial intelligence.
\newblock {\em arxiv preprint arxiv:1606.08514}, 2020.

\bibitem{HodgeJim2004}
Victoria Hodge and Jim Austin.
\newblock A survey of outlier detection methodologies.
\newblock {\em Artificial Intelligence Review}, 22(2):85--126, Oct 2004.

\bibitem{SharmaAzizanEtAl2021}
Apoorva Sharma, Navid Azizan, and Marco Pavone.
\newblock Sketching curvature for efficient out-of-distribution detection for
  deep neural networks.
\newblock In Cassio de~Campos and Marloes~H. Maathuis, editors, {\em
  Proceedings of the Thirty-Seventh Conference on Uncertainty in Artificial
  Intelligence}, volume 161 of {\em Proceedings of Machine Learning Research},
  pages 1958--1967. PMLR, 27--30 Jul 2021.

\bibitem{AminiSchwartingEtAl2020}
Alexander Amini, Wilko Schwarting, Ava Soleimany, and Daniela Rus.
\newblock Deep evidential regression.
\newblock In H.~Larochelle, M.~Ranzato, R.~Hadsell, M.F. Balcan, and H.~Lin,
  editors, {\em Advances in Neural Information Processing Systems}, volume~33,
  pages 14927--14937. Curran Associates, Inc., 2020.

\bibitem{LuoZhaoEtAl2021}
Rachel Luo, Shengjia Zhao, Jonathan Kuck, Boris Ivanovic, Silvio Savarese,
  Edward Schmerling, and Marco Pavone.
\newblock Sample-efficient safety assurances using conformal prediction.
\newblock In {\em Algorithmic Foundations of Robotics XV}, pages 149--169,
  Cham, 2023. Springer International Publishing.

\bibitem{FaridSnyderEtAl2022}
Alec Farid, David Snyder, Allen~Z. Ren, and Anirudha Majumdar.
\newblock Failure prediction with statistical guarantees for vision-based robot
  control.
\newblock {\em arxiv preprint arxiv:2202.05894}, 2022.

\bibitem{LeungSchmerlingEtAl2019}
K.~Leung, E.~Schmerling, M.~Zhang, M.~Chen, J.~Talbot, J.~C. Gerdes, and
  M.~Pavone.
\newblock On infusing reachability-based safety assurance within planning
  frameworks for human-robot vehicle interactions.
\newblock {\em {Int. Journal of Robotics Research}}, 39:1326--1345, 2020.

\bibitem{FisacAkametaluEtAl2019}
Jaime~F. Fisac, Anayo~K. Akametalu, Melanie~N. Zeilinger, Shahab Kaynama,
  Jeremy Gillula, and Claire~J. Tomlin.
\newblock A general safety framework for learning-based control in uncertain
  robotic systems.
\newblock {\em IEEE Transactions on Automatic Control}, 64(7):2737--2752, 2019.

\bibitem{BrunkeGreefEtAl2022}
Lukas Brunke, Melissa Greeff, Adam~W. Hall, Zhaocong Yuan, Siqi Zhou, Jacopo
  Panerati, and Angela~P. Schoellig.
\newblock Safe learning in robotics: From learning-based control to safe
  reinforcement learning.
\newblock {\em Annual Review of Control, Robotics, and Autonomous Systems},
  5(1):null, 2022.

\bibitem{FaridVeerEtAl2022}
Alec Farid, Sushant Veer, and Anirudha Majumdar.
\newblock Task-driven out-of-distribution detection with statistical guarantees
  for robot learning.
\newblock In {\em Proceedings of the 5th Conference on Robot Learning}, volume
  164 of {\em Proceedings of Machine Learning Research}, pages 970--980. PMLR,
  08--11 Nov 2022.

\bibitem{PodkopaevRamdasEtAl2022}
Aleksandr Podkopaev and Aaditya Ramdas.
\newblock Tracking the risk of a deployed model and detecting harmful
  distribution shifts.
\newblock In {\em International Conference on Learning Representations}, 2022.

\bibitem{RossGordon2011}
Stephane Ross, Geoffrey Gordon, and Drew Bagnell.
\newblock A reduction of imitation learning and structured prediction to
  no-regret online learning.
\newblock In {\em Proceedings of the Fourteenth International Conference on
  Artificial Intelligence and Statistics}, volume~15 of {\em Proceedings of
  Machine Learning Research}, pages 627--635, Fort Lauderdale, FL, USA, 11--13
  Apr 2011. PMLR.

\bibitem{LevineKumarEtAl2020}
Sergey Levine, Aviral Kumar, George Tucker, and Justin Fu.
\newblock Offline reinforcement learning: Tutorial, review, and perspectives on
  open problems.
\newblock {\em arxiv preprint arxiv:2005.01643}, 2020.

\bibitem{LesortLomonaco2020}
Timothée Lesort, Vincenzo Lomonaco, Andrei Stoian, Davide Maltoni, David
  Filliat, and Natalia Díaz-Rodríguez.
\newblock Continual learning for robotics: Definition, framework, learning
  strategies, opportunities and challenges.
\newblock {\em Information Fusion}, 58:52--68, 2020.

\bibitem{RichardsAzizanEtAl2021}
S.~M. Richards, N.~Azizan, J.-J.~E. Slotine, and M.~Pavone.
\newblock Adaptive-control-oriented meta-learning for nonlinear systems.
\newblock In {\em {Robotics: Science and Systems}}, Virtual, July 2021.

\bibitem{NagabandiClaveraEtAl2019}
Anusha Nagabandi, Ignasi Clavera, Simin Liu, Ronald~S. Fearing, Pieter Abbeel,
  Sergey Levine, and Chelsea Finn.
\newblock Learning to adapt in dynamic, real-world environments through
  meta-reinforcement learning.
\newblock In {\em 7th International Conference on Learning Representations,
  {ICLR} 2019, New Orleans, LA, USA, May 6-9, 2019}, 2019.

\bibitem{StokingGopnikEtAl2022}
Kaylene~Caswell Stocking, Alison Gopnik, and Claire Tomlin.
\newblock From robot learning to robot understanding: Leveraging causal
  graphical models for robotics.
\newblock In Aleksandra Faust, David Hsu, and Gerhard Neumann, editors, {\em
  Proceedings of the 5th Conference on Robot Learning}, volume 164 of {\em
  Proceedings of Machine Learning Research}, pages 1776--1781. PMLR, 08--11 Nov
  2022.

\bibitem{KirkZhangEtAl2021}
Robert Kirk, Amy Zhang, Edward Grefenstette, and Tim Rocktäschel.
\newblock A survey of generalisation in deep reinforcement learning.
\newblock {\em arxiv preprint arxiv:2111.09794}, 2021.

\bibitem{RusuColmenarejoEtAl2016}
Andrei~A. Rusu, Sergio~Gomez Colmenarejo, {\c{C}}aglar G{\"{u}}l{\c{c}}ehre,
  Guillaume Desjardins, James Kirkpatrick, Razvan Pascanu, Volodymyr Mnih,
  Koray Kavukcuoglu, and Raia Hadsell.
\newblock Policy distillation.
\newblock In Yoshua Bengio and Yann LeCun, editors, {\em 4th International
  Conference on Learning Representations, {ICLR} 2016, San Juan, Puerto Rico,
  May 2-4, 2016, Conference Track Proceedings}, 2016.

\bibitem{Settles2012}
Burr Settles.
\newblock Active learning.
\newblock {\em Synthesis Lectures on Artificial Intelligence and Machine
  Learning}, 6(1):1--114, 2012.

\bibitem{FilosTigas2020}
Angelos Filos, Panagiotis Tigas, Rowan McAllister, Nicholas Rhinehart, Sergey
  Levine, and Yarin Gal.
\newblock Can autonomous vehicles identify, recover from, and adapt to
  distribution shifts?
\newblock In {\em ICML}, ICML'20. JMLR.org, 2020.

\bibitem{RichterRoy2017}
Charles Richter and Nicholas Roy.
\newblock Safe visual navigation via deep learning and novelty detection.
\newblock In {\em Robotics: Systems and Science}, July 2017.

\end{thebibliography}
\end{document}